\pdfoutput=1

\documentclass[11pt]{article}

\usepackage[]{EACL2023}

\usepackage{times}
\usepackage{latexsym}
\usepackage{multirow}
\usepackage{booktabs}
\usepackage{graphicx} 

\usepackage[T1]{fontenc}

\usepackage[utf8]{inputenc}

\usepackage{microtype}

\usepackage{inconsolata}

\usepackage{todonotes}

\usepackage{tikz}
\usepackage{pgfplots}
\usepackage{tikz-dependency}
\usetikzlibrary{plotmarks}

\newcommand{\ACRO}[1]{\textsc{#1}}
\newcommand{\ARRAU}{\ACRO{arrau}}
\newcommand{\CRAFT}{\ACRO{craft}}
\newcommand{\DOMAINTRAIN}{\ACRO{Domain Only}}
\newcommand{\GAME}[1]{\textit{#1}}
\newcommand{\GWAP}{\ACRO{gwap}}
\newcommand{\LONGFORMER}{\ACRO{Longformer}}
\newcommand{\MPA}{\ACRO{mpa}}
\newcommand{\NLP}{\ACRO{nlp}}
\newcommand{\NP}{\ACRO{np}}
\newcommand{\ONTONOTES}{\ACRO{ontonotes}}
\newcommand{\PD}{\GAME{Phrase Detectives}}
\newcommand{\PDNOTANON}{\GAME{Phrase Detectives}}
\newcommand{\PDTHREE}{\GAME{Phrase Detectives} 3.0}
\newcommand{\PDGOLD}{\textsc{pd}\textsubscript{gold}}
\newcommand{\PDSILVER}{\textsc{pd}\textsubscript{silver}}
\newcommand{\PRECO}{\ACRO{preco}}
\newcommand{\PDTWOTRAIN}{\ACRO{Previous release}}
\newcommand{\PDCOMTRAIN}{\ACRO{Train Complete}}
\newcommand{\PDFULLTRAIN}{\ACRO{Train Full}}
\newcommand{\PDDEV}{\ACRO{Dev}}
\newcommand{\PDTEST}{\ACRO{Test}}
\newcommand{\ROBERTA}{\ACRO{Roberta}}

\title{Aggregating Crowdsourced and Automatic  Judgments to Scale Up a Corpus of Anaphoric Reference for Fiction and Wikipedia Texts}

\author{
\textbf{Juntao Yu}\textsuperscript{1},
\textbf{Silviu Paun}\textsuperscript{2}\thanks{Work was done prior to joining Amazon research.}, 
\textbf{Maris Camilleri}\textsuperscript{1},
\textbf{Paloma Carretero Garcia}\textsuperscript{1},\\
\textbf{Jon Chamberlain}\textsuperscript{1}, 
\textbf{Udo Kruschwitz }\textsuperscript{3} and 
\textbf{Massimo Poesio}\textsuperscript{4}\\
\textsuperscript{1}University of Essex, UK;
\textsuperscript{2}Amazon Research, Romania; \\
\textsuperscript{3}University of Regensburg, Germany;
\textsuperscript{4}Queen Mary University of London, UK;
\\
{\normalsize
\texttt{j.yu@essex.ac.uk; silviupn@amazon.com;  mcamil@essex.ac.uk;pcarre@essex.ac.uk;}}\\ {\normalsize\texttt{jchamb@essex.ac.uk; udo.kruschwitz@ur.de; m.poesio@qmul.ac.uk;}}
}

\begin{document}
\maketitle
\begin{abstract}
Although several datasets annotated for anaphoric reference / coreference exist, even the largest such datasets have limitations in term of size,  range of domains,  coverage of anaphoric phenomena, and size of documents included. 
Yet, the approaches proposed to scale up anaphoric annotation  haven't so far resulted in datasets overcoming these limitations. 
In this paper, we introduce a new release of a corpus for anaphoric reference labelled via a game-with-a-purpose.
This new release is comparable in size to the largest existing corpora for anaphoric reference due in part to substantial activity by the players, in part thanks to
the use of a new resolve-and-aggregate paradigm to `complete' markable annotations through the combination of an anaphoric resolver 
and an aggregation method for anaphoric reference. 
The proposed method could be adopted to greatly speed up annotation time in other projects involving games-with-a-purpose.
In addition, the corpus 
covers genres for which no comparable size datasets exist (Fiction and Wikipedia); 
it covers singletons and non-referring expressions; and 
it includes a substantial number of long documents ($>2K$ in length).
\end{abstract}

\section{Introduction}
\label{sec:introduction}

Many resources annotated for  anaphoric reference / coreference exist; but even the largest such datasets, such as {\ONTONOTES} \cite{pradhan2012conllst}, have a number of limitations. 
The largest  resources are still medium scale (e.g.,  {\ONTONOTES} \cite{pradhan2012conllst} is 1.5M tokens, as is {\CRAFT} \cite{cohen2017coreference}).  
They only cover a limited range of domains, primarily news (as in {\ONTONOTES}) and scientific articles (as in  {\CRAFT}), and models trained on these datasets have been shown not to generalize well to other domains \cite{xia&van-durme:EMNLP21}.\footnote{The largest existing corpus for English, the 10M words {\PRECO} \cite{chen-EtAl:2018:EMNLP}, consists of language learning texts, but the models trained on this genre have proven to have even worse performance on other domains.}
The range of anaphoric phenomena covered is also  narrow \cite{poesio-et-al:ana_book_corpora}.

Several proposals have been made to scale up anaphoric annotation in terms of size, range of domains, and phenomena covered  proposed, including automatic data augmentation \cite{emami-et-al:ACL19,gessler-et-al:LREC20,aloraini&poesio:CRAC21}, and crowdsourcing combined with active learning \cite{laws-et-al:NAACL2012,li(belinda)-et-al:ACL20,yuan-et-al:ACL22} or  through Games-With-A-Purpose \cite{Chamberlain08Phrase,hladka-et-al:LAW09,bos-et-al:HLA16,kicikoglu-et-al:GAMNLP19}.
However,  the largest existing  anaphoric corpora created using Games-With-A-Purpose  (e.g., \cite{poesio-et-al:NAACL19}) are still  smaller than the largest  resources created with traditional methods, and the corpora created using  data augmentation techniques are focused on  specific aspects of anaphoric reference. 
In order to use such approaches to create resources of the required scale in terms of size, variety and range of phenomena covered 
novel methods 
are required. 

The first contribution of this paper is the {\PDTHREE}  corpus of anaphoric reference annotated using a Game-With-A-Purpose. 
This corpus has a comparable size in tokens (1.37M) to the {\ONTONOTES} corpus \cite{pradhan2012conllst}, but twice the number of markables. 
Its annotation scheme also covers singletons and non-referring expressions; 
it is focused on two genres - fiction and Wikipedia articles - not covered in {\ONTONOTES}, and for which only much smaller datasets exist; 
and it includes a range of documents ranging from short to fairly long ($14K$ tokens)
enabling 
research on {\NLP} in 
long documents \cite{Beltagy2020Longformer}. 

The second contribution of the paper is a new iterative resolve-and-aggregate approach  developed to `complete' the corpus by combining crowdsourcing with automatic annotation. 
Only about 70\% of documents in the corpus were completely annotated by the players. 
The proposed method 
(i) uses an anaphoric resolver to automatically annotate  all mentions, including the few still unannotated; 
(ii) aggregates  the resulting judgments using a probabilistic aggregation method for anaphora, and 
(iii) uses the resulting expanded dataset to retrain the anaphoric resolver.
We show that the resolve-and-aggregate method results in models with 
higher 
accuracy than models trained using only the completely annotated data, or
the full corpus  not completed using the method.  

\section{Background}
\label{sec:background}

\paragraph{Anaphorically annotated corpora}

A number of anaphorically annotated datasets have now been released, 
covering a number of languages  
\cite{hinrichs-et-al:ACL05:frontiers,hendrickx-et-al:LREC08,recasens&marti:LRE10,pradhan2012conllst,%
landragin:16,nedoluzhko-et-al:LREC16,cohen2017coreference,chen-EtAl:2018:EMNLP,bamman-et-al:LREC20,uryupina-et-al:NLEJ,zeldes:book20} and 
turning anaphora / coreference in a very active area of research \cite{pradhan2012conllst,fernandes-et-al:CL14,wiseman-et-al:ACL15,
lee2017end,lee2018higher,yu-etal-2020-cluster,joshi(mandar)-et-al:TACL20:spanbert}.  
However, only a few of these resources 
are genuinely large 
in terms of markables 
\cite{pradhan2012conllst,cohen2017coreference}, and
most are focused on  news, with only a few 
corpora 
covering other genres  such as  
scientific articles (e.g., {\CRAFT} \cite{cohen2017coreference}),
fiction (e.g., LitBank \cite{bamman-et-al:LREC20} and {\PDNOTANON} \cite{poesio-et-al:NAACL19}),
and Wikipedia (e.g., WikiCoref \cite{ghaddar-langlais-2016-wikicoref} or again {\PDNOTANON} \cite{poesio-et-al:NAACL19}). 
Important genres such as dialogue are barely covered
\cite{muzerelle-et-al:LREC14,khosla-etal-2021-codi}. 
There is evidence that this concentration 
on a single genre, and  on  {\ONTONOTES} in particular, does not result in models that generalize well 
\cite{xia&van-durme:EMNLP21}. 

Existing resources are also  limited in terms of coverage. 
Most recent datasets are based on general purpose annotation schemes with a clear linguistic foundation, but especially the largest ones focus on the simplest cases of anaphora / coreference (e.g., singletons and non-referring expressions are not annotated in {\ONTONOTES}).
And the documents found in existing corpora tend to be short, with the exception of {\CRAFT}: e.g., average document length is 329 in {\PRECO}, 467 in {\ONTONOTES},  630 in {\ARRAU}, and 753 in {\PDNOTANON}. 

   
\paragraph{Scaling up anaphoric annotation}    

One approach  to scale up anaphoric reference annotation 
is using 
fully automatic methods to either annotate a dataset, such as \ACRO{amalgum} \cite{gessler-et-al:LREC20},
or create a benchmark from scratch,
such as \ACRO{knowref} \cite{emami-et-al:ACL19}.
While entirely automatic annotation  
may result in 
datasets of arbitrarily large size, 
such annotations 
cannot  expand current models' coverage to aspects of anaphoric reference do not already handle well. 
And creating from scratch  large-scale benchmarks  for specific issues
hasn't so far been shown to result in datasets reflecting the variety and richness of real texts. 


Crowdsourcing has emerged as the dominant paradigm for annotation in {\NLP} \cite{snow-et-al:08,poesio-et-al:handbook_anno_crowdsourcing} because of its reduced costs and increased speed in comparison with traditional annotation. 
But the costs for really large-scale annotation are still  prohibitive even for crowdsourcing \cite{Poesio13Phrase,poesio-et-al:handbook_anno_crowdsourcing}.
To address this issue,  a number of approaches have been developed to optimize the use of crowdsourcing for coreference annotation. 
In particular, active learning has been used to reduce the amount of annotation work needed \cite{laws-et-al:NAACL2012,li(belinda)-et-al:ACL20,yuan-et-al:ACL22}. 
Another issue is that anaphoric reference is a complex type of annotation whose most complex aspects require special quality control typically not available with microtask crowdsourcing. 

    
\paragraph{Games-With-A-Purpose}    

A form of crowdsourcing which has been widely used  to 
address the issues of cost and quality is Games-With-A-Purpose ({\GWAP}) \cite{ahn:06,cooper(s)-et-al:NATURE10_Foldit,lafourcade-et-al:GWAP_book_2015}.
%
Games-With-A-Purpose is the version of crowdsourcing where labelling is created through a game, so that the reward for the workers is in terms of enjoyment rather than financial--were proposed as a solution for large-scale data labelling. 
A number of {\GWAP}s were therefore developed for {\NLP}, including 
\GAME{Jeux de Mots} \cite{lafourcade:SNLP07,joubert-et-al:Games4NLP18},
{\PDNOTANON} \cite{Chamberlain08Phrase,Poesio13Phrase},
\GAME{OntoGalaxy} \cite{krause-et-al:HCOMP10},
the \GAME{Wordrobe} platform \cite{basile-et-al:LREC12},
\GAME{Dr Detective} \cite{dumitrache-et-al:13},
\GAME{Zombilingo} \cite{Fort14Creating},
\GAME{TileAttack!} \cite{madge-et-al:CHIPLAY17},
\GAME{Wormingo} \cite{kicikoglu-et-al:GAMNLP19},
\GAME{Name That Language!} \cite{cieri-et-al:Interspeech21} or 
\GAME{High School Superhero} \cite{bonetti&tonelli:HCINLP21}. 
%
{\GWAP}s for coreference 
include {\PDNOTANON} \cite{Chamberlain08Phrase,Poesio13Phrase}, 
the \GAME{Pointers} game in \GAME{WordRobe} \cite{bos-et-al:HLA16} and
\GAME{Wormingo} \cite{kicikoglu-et-al:GAMNLP19},
all deployed, 
and \GAME{PlayCoref} \cite{hladka-et-al:LAW09}, proposed but not tested.

However, whereas truly successful {\GWAP}s such as \ACRO{foldit} have been developed in other areas of science
\cite{cooper(s)-et-al:NATURE10_Foldit}, 
even the most successful {\GWAP}s for {\NLP}  only collected moderate amounts of data \cite{poesio-et-al:NAACL19,joubert-et-al:Games4NLP18}. 
In part, this is because the games 
used to actually collect {\NLP} labels
aren't very entertaining, leading to
efforts  to develop  engaging designs such as 
\cite{jurgens&navigli:TACL14,dziedzic&wlodarczyk:GAMES4NLP17,madge-et-al:CHIPLAY19}. 

An interesting solution to this issue 
was proposed although not fully developed for  
\GAME{Wordrobe}  \cite{bos-et-al:HLA16}. 
This solution is a hybrid between automatic annotation and crowdsourcing: 
a combination of crowd and automatically computed judgments is aggregated to ensure that every item has at least one label.  
This solution wasn't  properly tested in \GAME{Wordrobe}, which only collected very few judgments and for a small corpus; and anyway the approach followed could not be applied to anaphora/coreference, due to the lack of a suitable aggregation mechanism for anaphora/coreference. 
In this paper we propose a method for aggregating crowd and automatic judgments inspired by this idea, but using an aggregation method for anaphora,  and truly tested on a dataset containing a very large number  of anaphoric judgments.



\section{Phrase Detectives}
\label{sec:game}


The {\PD} Game-With-A-Purpose 
\cite{Chamberlain08Phrase,Poesio13Phrase,chamberlain:thesis,poesio-et-al:NAACL19}
was designed to collect multiple  judgments about anaphoric reference.

\paragraph{Game design} {\PD} doesn't follow the design 
of some of the original von Ahn games \cite{ahn&dabbish:08}, in that 
it is a one-person game, and not timed; both competition and timing were found to have orthogonal effects on the quality of the annotation \cite{chamberlain:thesis}. 
Points are used as the main incentive, with weekly and  monthly boards being displayed. 

Players play  two different games:
one aiming at labelling new data, the other at validating judgments expressed by the other players. 
In the annotation game, \GAME{Name the Culprit},
the player 
provides an anaphoric judgment
about a highlighted markable 
(the possible judgments  according to the annotation scheme are discussed next). 
If different participants 
enter different interpretations for a markable then each interpretation is presented to other 
participants 
in the validation game, \GAME{Detectives Conference}, 
in which the 
participants 
have to agree or disagree with the interpretation.

Every item is annotated by at least 8 players (20 on average), and each distinct interpretation is validated by at least four players. 
Players get  points for each label they produce, but especially when their interpretation is agreed upon by other players, thus rewarding accuracy. 
Initially, players play against gold data,
and are periodically evaluated against the gold; when they achieve a sufficient level of accuracy, they start seeing 
incompletely annotated data.  
Extensive  analyses of the data suggest that although there is a great number of noisy judgments, this simple training and validation method delivers extremely accurate aggregated labels 
\cite{Poesio13Phrase,chamberlain:thesis,poesio-et-al:NAACL19}

\begin{table*}
\centering
\small
\begin{tabular}{llcccc}
\toprule
\textbf{Type} & \textbf{Example} & \ONTONOTES & \PRECO & \ARRAU & {\PD} \\\midrule
predicative {\NP}s & [John] is \underline{a teacher} & Pred & Coref & Pred & Pred \\
   & [John, \underline{a teacher}] & & & & \\
singletons & & No & Yes & Yes & Yes \\
expletives & \underline{It}'s five o'clock & No & No & Yes & Yes \\
split antecedent plurals & [John] met [Mary] & No & No & Yes & Yes\\
 & and \underline{they} ... & & & & \\
generic mentions & [Parents] are usually busy. & 
 Only with  & Yes & Yes & Yes\\
    & \underline{Parents} should get involved & 
  pronouns   &    &     &     \\
event anaphora & Sales [grew] 10\%. & Yes & No & Yes & No \\
  & \underline{This growth} is exciting & & & 
  & \\
ambiguity & Hook up [the engine] & No & No & Explicit & Implicit \\
    & to [the boxcar] & & & & \\
    & and send \underline{it} to Avon \\
\bottomrule
\end{tabular}

\caption{Comparison between the annotation schemes in {\ONTONOTES}, {\PRECO}, {\ARRAU} and the {\PD} corpus}
\label{tab:anno_scheme_comparison}
\end{table*}

\paragraph{Annotation Scheme} The annotation scheme used in {\PD} is a simplified version of the {\ARRAU} annotation scheme \cite{uryupina-et-al:NLEJ},  covering all the main aspects of anaphoric annotation, including 
the distinction between referring and non-referring expressions 
(all noun phrases are annotated as either referring or non-referring, with two types of non-referring expressions being annotated: expletives and predicative {\NP}s);
the distinction between discourse-new and discourse-old referring expressions \cite{prince:92};
and the annotation of all types of identity reference (including split antecedent plural anaphora). 
Only 
the most complex types of anaphoric reference (bridging references and discourse deixis) are not annotated. 
The main differences between the annotation scheme used in {\PD} and those used in {\ARRAU}, {\ONTONOTES}, and {\PRECO} 
are summarized in Table \ref{tab:anno_scheme_comparison}, modelled on a similar table in \cite{chen-EtAl:2018:EMNLP}.
In the {\PD} 
corpus  
predication and coreference are 
clearly distinguished, 
as in {\ONTONOTES} and  {\ARRAU} but unlike in {\PRECO}. 
Singletons are considered 
markables.
Expletives and 
split antecedent plurals are marked, unlike in either {\ONTONOTES} or {\PRECO}.

Possibly the most distinctive feature of the annotation scheme 
is that disagreements among annotators are preserved, encoding a form of implicit ambiguity as opposed to the explicit ambiguity annotated in {\ARRAU}. 

The {\PDDEV} and {\PDTEST} subsets of the corpus (see next Section) were annotated according to the full {\ARRAU} scheme. 

\paragraph{Preliminary player statistics} At the time of writing (11th October, 2022), 
61,391 players 
have registered,
of which more than 4,000 graduated to 
being allowed to provide judgments on partially labelled data. 
So far, the players have provided about 3.7M annotations and 1.7M validations, for a total of over 5.4M judgments. 

\paragraph{Speed of annotation} Over the course of the project, the games has been collecting an average of 385,000 judgments a year, i.e., slightly over 1,000 judgments per day, every day. 
While this is an impressive number of judgments, it only translates in an average of around 10,000 new completely annotated markables per year, or 20 new completely annotated documents, for an average of 30,000 extra words. 
(Progress was faster 
in the early years of the project, when all short documents were annotated; but as discussed in the next Section, the corpus also contains a number of fairly long texts -- these are the ones being annotated now.)
The project discussed in this paper was motivated by the simple calculation that at this speed, it would take us 40 years to completely annotate all the documents  already in  the corpus, and 300 years to completely annotated a corpus of 10M words.


\section{Characteristics of the corpus}


The {\PDTHREE} corpus includes all the 805 documents 
originally uploaded 
in the game. 
In this Section we highlight the 
main characteristics of the texts in this release, summarized in Table \ref{tab:PD3_summary}.
For comparison, we include  in the Appendix a short description of 
the previous 
release of the {\PD} corpus, {\PD} 2, released in 2019 \cite{poesio-et-al:NAACL19}.

\begin{table}[]
    \centering
    \resizebox{0.84\linewidth}{!}{
    \begin{tabular}{ccccc}
    \toprule
              & & Docs & Tokens & Markables 
              \\\midrule
        \multirow{4}{*}{\shortstack{\ACRO{Train}\\ \ACRO{Complete}}}   
       & Gutenberg & 154 & 181142 & 48329 (29527) \\
& Wikipedia & 359 & 244770 & 65050 (21803) \\
& Other & 2 & 7294 & 2126 (1347) \\
& Subtotal & 515 & 433206 & 115505 (52677) \\
       \midrule
     \multirow{4}{*}{\shortstack{\ACRO{Train}\\ \ACRO{Full}}}   
       & Gutenberg & 194 & 372001 & 102354 (57387) \\
& Wikipedia & 544 & 931752 & 258560 (92465) \\
& Other & 2 & 7294 & 2128 (1347) \\
& Subtotal & 740 & 1311047 & 363042 (151199) \\
       \midrule
    \multirow{4}{*}{\PDDEV}   
       & Gutenberg & 5 & 7536 & 2133 (1494) \\
& Wikipedia & 35 & 15287 & 4423 (1669) \\
& Other & 5 & 989 & 331 (126) \\
& Subtotal & 45 & 23812 & 6887 (3289) \\
       \midrule
    \multirow{3}{*}{\PDTEST}   
       & Gutenberg & 7 & 20646 & 5925 (3332) \\
& Wikipedia & 13 & 22998 & 7704 (3876) \\
& Subtotal & 20 & 43644 & 13629 (7208) \\
       \midrule
    \multirow{4}{*}{All}
    & Gutenberg & 206 & 400183 & 110412 (62213) \\
& Wikipedia & 592 & 970037 & 270687 (98010) \\
& Other & 7 & 8283 & 2459 (1473) \\
       & Total    & 805 & 1378503 & 383558 (161696)\\
       \bottomrule
    \end{tabular}}
    \caption{Summary 
    of the current release. 
    In parentheses 
    the 
    number of markables that are non-singletons. 
    }
    \label{tab:PD3_summary}
\end{table}


\paragraph{The new release}
The new release of the corpus, {\PDTHREE}, is more than three times larger than the previous release of the {\PD} corpus described in the previous section in terms of the number of tokens (1.4M) and markables (383K). 
(See `All' row in Table \ref{tab:PD3_summary}.)
This makes the {\PDTHREE} corpus comparable in 
the number of 
tokens 
to  {\ONTONOTES}, 
but double the size of {\ONTONOTES}
in terms of markables,
partly due to the singletons and non-referring expressions being included. 
72\% of the documents were completely annotated by the game players (580 out of 805 documents), and  the near totality of mentions have at least one annotation from the crowd (99.4\%). 

\paragraph{Genres}
The corpus covers mainly two genres. 
The Gutenberg domain consists of fiction texts from the Gutenberg Project:
in part fairy tales (e.g., \textit{Alice in Wonderland}, 
Grimm brothers stories), in part classics (e.g., 
Sherlock Holmes 
stories). 
At 400K tokens, it is twice the size of the largest existing fiction corpus \cite{bamman-et-al:LREC20}.
The Wikipedia domain consists of primarily the `Wikipedia Unusual' documents.
This 
subset is  1M tokens in size, substantially larger than  WikiCoref  (60K tokens) \cite{ghaddar-langlais-2016-wikicoref}.


\paragraph{Organization}
The corpus is split into train, development, and test sets, where the development and test sets are annotated by human experts (see below) and the training set is aggregated using the {\MPA} anaphoric annotation model \cite{paun-et-al:EMNLP18} 
as 
described in Section \ref{section-resolve-and-aggregate}. 
But crucially, 
two versions of the training set exists. 

{\PDCOMTRAIN} is like the training sets released in previous versions of the corpus, in that it
consists of documents that were completely annotated by the players: i.e., all markables in the documents have more than 8 judgments, and all interpretations have more than 4 validations. 

The second training set, \PDFULLTRAIN,
additionally includes documents that have not yet been `completely' annotated by the players. 
These documents are considerably longer, and as a result it is harder to have them completely annotated.
So a state-of-the-art coreference model for this annotation scheme  \cite{yu-etal-2020-cluster} was used in the resolve-and-aggregate setting discussed in Section \ref{section-resolve-and-aggregate} to augment the existing annotations by ensuring that every markable had at least one label, which would then be aggregated with the others.
{\PDFULLTRAIN} is three times larger than {\PDCOMTRAIN}, both in the number of tokens and of markables. 


\paragraph{A New Gold}
The test set from the previous release of the corpus, 
45 documents, is now available as {\PDDEV}.
{\PDDEV}  was fully revised by human experts for this release to correct previous labelling mistakes, and is now  annotated according to the full {\ARRAU} guidelines, including ambiguity annotation, bridging references, and discourse deixis. 
In addition, a brand new {\PDTEST} set of 20 documents was also created, balanced between the two domains,  double in size compared to the old test set, and also annotated according to the full {\ARRAU} guidelines.

\paragraph{Domain specific training}
With the new release the corpus is also large enough to be used separately for domain-specific research. 
We demonstrate  in Section \ref{section-results} that models trained on the domain-specific portion of the training set can achieve comparable results to those trained on {\PDFULLTRAIN}. The results indicate that the domain-specific training data can be sufficient to be used separately for dedicated research in target domains.

\paragraph{Long and short documents}
An important characteristic of the corpus is that it was designed to  contain both short documents (< 2K tokens) and long ones. 
34.5\% of the documents are longer than 2K tokens, and the longest document reaches 14K tokens. 
(In contrast, in  {\ONTONOTES}  only 0.4\% of the documents have more than 2K tokens.)
This makes our corpus a suitable resource for research on long-distance anaphora and on long document training. 
To this end, we use our dataset to replicate  the experiments by \newcite{Beltagy2020Longformer} comparing the {\LONGFORMER} model with the {\ROBERTA} model. 
In the original paper, which used the {\ONTONOTES} corpus, no obvious differences were found between the two models,
partly due to the lack of long documents.
We discuss these experiments in Section \ref{section-long-docs-results}.
The only other corpus that we are aware of with a large portion of long documents is the {\CRAFT} corpus \cite{cohen2017coreference}, which is however focused  on biomedical texts. 

\section{Resolve-and-Aggregate}
\label{section-resolve-and-aggregate}



\paragraph{The challenge}
To create a reliable corpus using crowdsourcing, 
multiple judgments are required to ensure a good coverage of correct answers, together with sufficient evidence to enable an accurate aggregation method \cite{paun-et-al:TACL2018, paun-artstein-poesio-book} to  distill the correct answers from the noisy ones.
The problem of collecting such large number of judgments is even more serious for long documents. 
Annotating all anaphoric relations in long documents is  challenging, partly due to the amount of time needed to complete the task, but also because   the great number of entities makes it  difficult for  annotators to keep track of all the coreference chains. 
And indeed, 
the short documents in our corpus were completed much faster than the longer documents: the average 
length of the incomplete documents is 4K tokens, 
whereas for the complete documents is 850 tokens.
Thus in our corpus 
the rate at which judgments are collected from  players, while substantial (over 1,000 judgments per day) is not sufficient to extract reliable labels in a reasonable amount of time, as discussed in Section \ref{sec:game}.

\paragraph{Possible solutions}
Clearly, part of the solution is to develop more engaging games, thus able to attract more players and keep them playing for longer \cite{ahn&dabbish:08,jurgens&navigli:TACL14,madge-et-al:CHIPLAY19,kicikoglu-et-al:GAMNLP19}. 
A second ingredient is to use active learning-like approaches to minimize the number of labels required to complete the annotation \cite{laws-et-al:NAACL2012,li(belinda)-et-al:ACL20,yuan-et-al:ACL22}.
A number of proposals have been made in these two directions, and we are carrying out research in these  areas as well, reported elsewhere. 
In this work however we investigate an approach that to our knowledge has been  much less studied: 
combining crowdsourcing with automatic labelling.
Specifically, we propose a new resolve-and-aggregate method that iteratively makes use of a coreference resolver to enhance the collected annotations. 
The approach is inspired, apart from  \GAME{Wordrobe} \cite{bos-et-al:HLA16},
by previous work on Bayesian combination of classifiers \cite{bcc-2012} which allows for aggregating predictions from classifiers and humans together with the help of a probabilistic annotation model. Both the iterative use of the coreference resolver and the application domain of the annotation model are however novel to this paper. 


\paragraph{The coreference resolver}
As a coreference resolver, we use the system by  \citet{yu-etal-2020-cluster} which, to the best of our knowledge,  is the only modern coreference resolver that also predicts singletons and non-referring expressions, both of which are need to be 
annotated in our corpus. 
The system is an extension of  \citep{lee2017end,lee2018higher}, replacing their mention-ranking algorithm with a cluster-ranking algorithm to build the entity clusters incrementally. 
The system uses BERT \citep{devlin-etal-2019-bert} for pre-trained contextual embeddings instead of the Elmo embeddings \citep{peters2018elmo} used in \citep{lee2018higher}. 


\paragraph{Aggregation}
Standard aggregation methods for classification labels such as the \citep{dawid&skene:79} model are not appropriate for coreference labels, whose class space is not fixed but depends on the document mentions.
However, an aggregation model for coreference judgments 
is now available, 
the mention-pair annotation model ({\MPA})  \citep{paun-et-al:EMNLP18}.
We used {\MPA} to aggregate judgments by players and by the coreference resolver. 
{\MPA} 
can capture the accuracy and bias of the players, and of the coreference resolver, respectively, and  adjust the aggregated labels accordingly. 

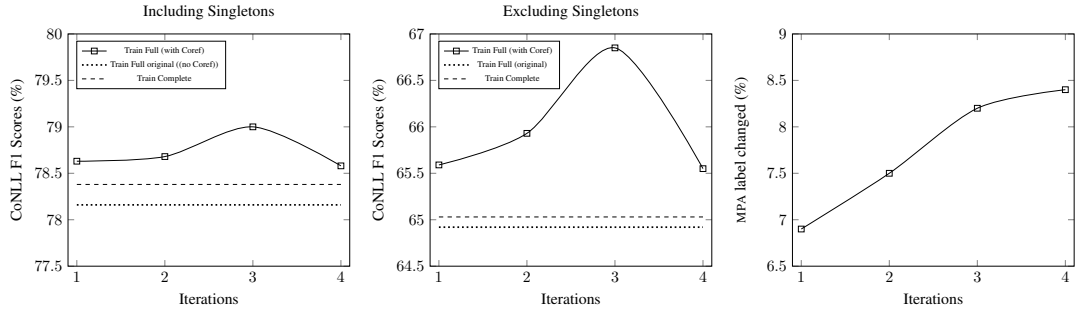
\begin{figure*}[t]
	\centering
    \resizebox{0.9\textwidth}{!}{
    \begin{tabular}{ccc}
         \begin{tikzpicture}
		\pgfplotsset{
			xmin=0.9,xmax=4.1,
			xtick={1,2,...,4},
			ymin=77.5,ymax=80,
			xlabel= Iterations,
			ylabel=CoNLL F1 Scores (\%),
			title = Including Singletons,
			legend pos=north west
		}
		\begin{axis}
		\addplot[smooth,mark=square]
		coordinates{(1,78.63) (2,78.68) (3,79.00) (4,78.58)};
		\addlegendentry{\tiny Train Full (with Coref)}
		
		\addplot[smooth, very thick, dotted] 
		coordinates{(1,78.16) (4,78.16)};
		\addlegendentry{\tiny Train Full original ((no Coref))}
		
		\addplot[smooth,dashed] 
		coordinates{(1,78.38) (4,78.38)};
		\addlegendentry{\tiny Train Complete}
		
		\end{axis}
		\end{tikzpicture}
		
		&
		
		\begin{tikzpicture}
		\pgfplotsset{
			xmin=0.9,xmax=4.1,
			xtick={1,2,...,4},
			ymin=64.5,ymax=67,
			xlabel= Iterations,
			ylabel=CoNLL F1 Scores (\%),
			title= Excluding Singletons,
			legend pos=north west
		}
		\begin{axis}

		\addplot[smooth,mark=square, color=black]
		coordinates{(1,65.59) (2,65.93) (3,66.85) (4,65.55)};
		\addlegendentry{\tiny Train Full (with Coref)}
		
		\addplot[smooth, very thick, dotted] 
		coordinates{(1,64.92) (4,64.92)};
		\addlegendentry{\tiny Train Full (original)}
		
		\addplot[smooth,dashed] 
		coordinates{(1,65.03) (4,65.03)};
		\addlegendentry{\tiny Train Complete}
		\end{axis}
		
		\end{tikzpicture}
		
		&
		
		\begin{tikzpicture}
		\pgfplotsset{
			xmin=0.9,xmax=4.1,
			xtick={1,2,...,4},
			ymin=6.5,ymax=9,
			xlabel= Iterations,
			ylabel= {\MPA} label changed (\%),
			legend pos=north west
		}
		\begin{axis}

		\addplot[smooth,mark=square]
		coordinates{ (1,6.9) (2,7.5) (3,8.2) (4,8.4)};
		
		\end{axis}
		
		\end{tikzpicture}
         
    \end{tabular}

	}
	\caption{\label{figure:dev-results-mpa-label} Left and Middle: The CoNLL scores for  \newcite{yu-etal-2020-cluster}  trained on different training sets and tested on the {\PDDEV} set in gold mention setting. Right: The percentage of {\MPA} labels changed by using the additional judgments from the \newcite{yu-etal-2020-cluster} system in different iterations. }
\end{figure*}


		
		

\paragraph{Resolve-and-aggregate} resolve-and-aggregate  is an iterative procedure which relies on the {\MPA} aggregation model to label the corpus, which is in turn used to retrain the coreference resolver to get better system predictions. 
More specifically, in the first step of the procedure we aggregate the players' annotations from the complete documents and build an initial training set, {\PDCOMTRAIN}. 
Then, we train the coreference resolver on this set, but in a gold mention setting to mimic the players who focus only on the resolution task. Having trained the system, we then use it to get predictions for the entire dataset. 
The resolver can be seen as a player who played all the documents in the corpus. 
Next, all the players' annotations and the system's predictions are aggregated using {\MPA}, and an initial version of the entire corpus, {\PDFULLTRAIN}, is built as a result. 
This  procedure is repeated, taking {\PDFULLTRAIN} as input and creating a new version every time. 
With each new version, the {\MPA}-aggregated labels get refined, leading in turn to better predictions from the coreference resolver. The procedure is repeated until the performance of the resolver plateaus. 
The final version of the corpus contains the {\MPA}-aggregated labels of the players' annotations and the system's best predictions. 
We show in the next Section that this approach results in substantial improvements in the quality of the labels produced by the coreference resolver, which translate in more accurate labels for the items not
fully annotated by the players. 

\section{Resolving-and-Aggregating results}
\label{section-results}

\paragraph{Experiment Setting}
For our experiments, we report the CoNLL F1 scores as defined in \cite{pradhan2012conllst} in both singleton included and excluded settings, as well as non-referring F1 scores for non-referring expressions. 
We use the Universal Anaphora (UA) Scorer \cite{yu-EtAl:2022:LREC2} 
that 
reports all the necessary 
scores. 

We trained the \newcite{yu-etal-2020-cluster} system using most of its default settings. The only exception is that we always use the full context of the documents for training instead of choosing a random 1K tokens as done in \newcite{yu-etal-2020-cluster}. The default setting gives priority to the short documents as for each epoch, the full context of the short documents is always used, whereas for long documents only part of the documents is used. 

We establish three baselines,
all using 
the same system  \newcite{yu-etal-2020-cluster}  with the same settings, 
but trained with different 
training sets. 
The first baseline is trained on the {\PDTWOTRAIN}.
The second baseline is trained on  {\PDCOMTRAIN} (complete documents aggregated by  {\MPA}  without resolve-and-aggregate). 
The third baseline is trained on 
{\PDFULLTRAIN} aggregated by  {\MPA} but without annotations from the coreference resolver.

\subsection{Testing on the Development set}
We first trained the system using the gold mention settings to improve the quality of the corpus. 
We used the baseline trained on 
{\PDCOMTRAIN} 
to annotate the full corpus, then assigned labels to all the mentions by aggregating  player and system annotations using {\MPA}. 
We then trained a new model by using the full corpus ({\PDFULLTRAIN} (with Coref)) and doing  resolve-and-aggregate between the system and {\MPA} in iterations until the system performance stopped improving. 

The first key result is that, 
the system trained with  {\PDFULLTRAIN} (with Coref) always outperforms the baseline trained on the {\PDCOMTRAIN} (see Figure \ref{figure:dev-results-mpa-label}). 
The improvements on the singletons excluded setting are larger than those in the singletons included setting; this makes sense as all the models use gold mentions, hence the performance with singletons is inflated by the gold mentions. The system achieved the best performance on the third iteration with CoNLL F1 scores of 79\% and 66.9\% for singletons included and excluded settings respectively. This is 0.6\% and 1.9\% higher than the {\PDCOMTRAIN} baseline. 

What is especially interesting is that the improvement is not just a matter of {\PDFULLTRAIN} being larger than {\PDCOMTRAIN}: running the coreference resolver helps substantially. The system trained on  {\PDFULLTRAIN} original (i.e., without any automatic labels) is slightly worse than the {\PDCOMTRAIN} baseline, despite using the additional training data. One explanation would be that {\MPA}'s performance 
is affected 
by the lower 
number of 
judgments collected in the incomplete documents: the correct answer might not appear in the players annotations, or the players producing the annotations might not considered sufficiently reliable. 

\begin{table}[t]
\centering
\resizebox{0.9\linewidth}{!}{
\begin{tabular}{l c c c}
\toprule
& \multicolumn{2}{c}{CoNLL Avg. F1}     &     \\ 
Train data&Sing. (inc)&Sing. (exc)&NR F1\\\midrule
\PDTWOTRAIN  &65.5 & 53.6 & 36.8  \\ 
\PDCOMTRAIN &66.1 & 54.7 & 39.4 \\
\PDFULLTRAIN (original)&64.9 & 52.9 & 35.5\\
\PDFULLTRAIN (with Coref)& \bf 66.8 & \bf 56.1 & \bf 40.1\\
\bottomrule
\end{tabular}
}
\caption{The CoNLL and non-referring scores for \newcite{yu-etal-2020-cluster} system trained on different training sets and tested on the {\PDTEST} set in predicted mention setting.}
\label{tab:conll-non-referring-results-pred-mention}
\end{table}

To quantify the contribution of the automatic coreference resolver, 
we 
calculate the percentage of {\MPA} labels flipped due to the additional system annotations. 
We compare the labels of the {\PDFULLTRAIN} (with Coref) in different iterations with the  {\PDFULLTRAIN} (original) labels. We find that in the first iteration, 7\% of the {\MPA} labels (26K) were 
changed (see Figure \ref{figure:dev-results-mpa-label}). The percentages increased sharply until iteration 3 to 8.2\% (31K) but slowed down for iteration 4.  
This 
might explain 
why performance starts dropping 
in the 4th iteration. 

{\MPA} works very well when the number of judgments is high, 
but 
performance might be affected when there are not enough annotations, e.g. for the incomplete documents. 
We suspected  {\MPA} might benefit more from  system annotations when the document is incomplete.  To assess our hypothesis, we took a closer look at {\MPA} labels from our best iteration. We split the documents into two classes,  complete and incomplete, according to our complete criterion (i.e., a document is considered complete when every markable has been annotated by at least 8 players, and each distinct interpretation has been validated by at least 4 players) and calculate a separate score for each class. 
We find that for the complete document only 3.3\% of the {\MPA} labels are changed as a result of the additional system annotations;  in contrast, 10.8\% of  {\MPA} labels are changed in the incomplete documents.

To assess the quality of these label changes, we checked the different {\MPA} labels between iteration 3 and the original on the {\PDDEV} set. Since all documents from the {\PDDEV} set are complete documents, out of 7K mentions, only 201 
have a different label. The {\PDFULLTRAIN} (original) gets 70 of the labels correct with an accuracy of 34.8\%, whereas after the 3rd iteration of resolve-and-aggregate, the number increased to 125 (62.2\% accuracy). Although the sample is not large, it still gives a clear picture that even for complete documents the system annotations can improve the quality of the corpus.

\subsection{Evaluation on the Test set}
\label{section-test-results}
After finding the best setting as discussed in the previous Section, we evaluated 
the impact of resolve-and-aggregate 
on the {\PDTEST} set in the more realistic predicted mention setting. 
As 
shown in 
Table \ref{tab:conll-non-referring-results-pred-mention}, our best model (trained on {\PDFULLTRAIN (with Coref)}) outperforms all the baselines in both singletons included and excluded settings. Of the baselines trained on the complete documents only, the {\PDCOMTRAIN} baseline works better than the {\PDTWOTRAIN} baseline, 
most likely because the training set is larger while the quality of the annotation remains the same.
But again, when training with the additional 
incomplete documents ({\PDFULLTRAIN} (original without Coref)), the performance dropped substantially
by 1\%-2\% when compared with the {\PDCOMTRAIN} baseline. This again highlights the importance of combining automatic and crowd annotations via  resolve-and-aggregate: the model trained on this corpus significantly outperforms the {\PDFULLTRAIN} (original) baseline by up to 3.2\%. 
The story remains the same 
for 
the models' performance on non-referring expressions, illustrated in 
Table \ref{tab:conll-non-referring-results-pred-mention}:
again, the model trained on {\PDFULLTRAIN} (with Coref) is top of the list.

\subsection{Annotation speed-up}

The results showed in the previous Sections show that using automatic annotations turns the incomplete documents into documents whose quality is enough to result in improved performance when training a coreference resolver, 
speeding up annotation. 
In this section, we try to estimate the amount of time potentially saved by the proposed method. For the complete documents, we have on average 20 judgements (annotations and validations) per markable, which seems sufficient to  
ensure 
the quality of the corpus, 
if not perhaps necessary. 
For incomplete documents, the average number of judgements is currently 7.7. 
If we do need 20 judgements to achieve the same quality as the complete documents, we still need to collect on average 12.3 more judgements for every markable.  
Multiplied by the number of markables in the incomplete documents (250K), this means we would need 3M more judgements to complete all documents in the game. 
In the last five years, we have been averaging 334K judgements per year, which means if we proceed at the current speed, we need another 9 years before we can release this corpus. In other words, the resolve-and-aggregate method significantly speeds up the annotation process.


\begin{table}[t]
\centering
\resizebox{0.85\linewidth}{!}{
\begin{tabular}{l c c c}
\toprule
& \multicolumn{2}{c}{CoNLL Avg. F1}     &     \\ 
Train data&Sing. (inc)&Sing. (exc)&NR F1\\\midrule
\multicolumn{4}{c}{Gutenberg}\\\midrule
\DOMAINTRAIN  & 70.4 & 61.8 & 43.8 \\ 
{\PDFULLTRAIN} (with Coref) & \bf 71.5 & \bf 62.1 & \bf 44.9\\ \midrule
\multicolumn{4}{c}{Wikipedia}\\\midrule
\DOMAINTRAIN  & 61.9 & \bf 50.9 & \bf 36.1\\ 
{\PDFULLTRAIN} (with Coref) &\bf 62.3 & 50.6 & 36.0 \\ 
\bottomrule
\end{tabular}
}
\caption{The CoNLL and non-referring scores for the system trained on different training sets and tested on the {\PDTEST} set of different domains using predicted mention. }
\label{tab:conll-non-referring-results-pred-mention-domain}
\end{table}

\subsection{Domain-specific Training}
\label{sec:domain-specific}

Thanks to 
the resolve-and-aggregate method, 
this new release 
gives us  datasets of a reasonable size for both the Gutenberg (fiction) and Wikipedia domains. 
We 
evaluated  system performance on the domain-specific portion - e.g., for Fiction we trained our model on the Gutenberg section of  {\PDFULLTRAIN} (with Coref) and tested it on the Gutenberg section of the {\PDTEST}. 
We then compared the performance of these domain-specific models with that of 
the best system trained on the entire corpus. 
As shown in  
Table \ref{tab:conll-non-referring-results-pred-mention-domain},
the {\DOMAINTRAIN} systems trained on the domain-specific subsections of the corpus  achieve scores close to the system trained on the full corpus. 
This suggests each domain-specific part of the corpus is sufficiently large to be used for domain-specific research. 

\begin{table}[t]
\centering
\resizebox{0.8\linewidth}{!}{
\begin{tabular}{l c c c}
\toprule
Model&Short Doc&Long Doc&All Doc\\\midrule
\LONGFORMER  & \bf 61.0 & \bf 67.2 & \bf 64.7 \\ 
\ROBERTA & 60.1 & 65.2& 63.1\\ 
\bottomrule
\end{tabular}
}
\caption{The CoNLL scores (exclude singltons) for {\LONGFORMER} and {\ROBERTA} trained on {\PDFULLTRAIN} and tested on the {\PDTEST} set using gold mentions.}
\label{tab:test-results-pred-mention-longformer}
\end{table}

\subsection{Long and short documents}
\label{section-long-docs-results}

As stated earlier, one of the emerging challenges for  research on anaphora  (and {\NLP} in general)
are 
longer documents (>2K tokens). 
Our corpus is unusual in that it includes a large number of documents  more  than 2K in length, with the longest document containing 14K tokens. 
{\PDTEST}  also balances short (55\%) and long (45\%) documents. 

To test that the corpus can support research on anaphora in long documents, 
we used it to replicate the comparison in \cite{Beltagy2020Longformer} between their new model designed specifically for longer documents, the {\LONGFORMER}, with {\ROBERTA} \cite{liu2019roberta}. 
In that paper, the {\LONGFORMER} is compared  with {\ROBERTA} on the {\ONTONOTES} corpus,
without however finding a clear difference between the two systems. 
We suspected 
this might be because
{\ONTONOTES} does not contain enough long documents to observe 
improvements. 
We replicated the experiments by \citeauthor{Beltagy2020Longformer} with our corpus, and report the CoNLL F1 score on full {\PDTEST} as well as separate scores for long/short documents. (Since neither system predicts singletons and non-referring expressions, we report the CoNLL F1 scores in the singleton excluded setting.) 
We evaluated the systems with the gold mentions so that the system's performance will not be affected by  mention detection. 

Table \ref{tab:test-results-pred-mention-longformer} shows the results for both systems on different test set. 
The {\LONGFORMER}  works better on all  test sets, but with a much larger gain over  {\ROBERTA} on long documents:  the improvement over {\ROBERTA}  is 0.9\% and 2\% 
on  short and long documents respectively. 
This finding confirms that long documents benefit more from the {\LONGFORMER} architecture, while also showing that our corpus can be used to differentiate systems designed to perform on long documents.



\section{Conclusion}


This research makes two main contributions. 
First of all, we proposed an iterative method for speeding up anaphoric annotation via 
{\GWAP}s 
by combining crowdsourced data with labels produced by an automatic coreference resolver, and aggregating the labels using a probabilistic annotation method; and showed that the resulting extension leads to quantifiable improvements in model performance. 
The method can be easily extended to other types of annotation. 
Second, we introduced a new corpus for anaphoric reference which, thanks to the use of resolve-and-aggregate, is of a comparable size to {\ONTONOTES} in terms of tokens, but twice the size in terms of markables; it contains two substantial datasets for genres not covered in {\ONTONOTES}; and it includes both short and long documents. 
The corpus will be made freely available with all judgements. 

\section{Limitations}
\label{sec:limitations}

The main limitation of this work is that the  new release is still only twice the size of {\ONTONOTES} in terms of markables. 
In ongoing work, we are developing a new platform to label a corpus twenty times the size of the current release.
The new platform combines more engaging games with active-learning like methods for allocating work to players more efficiently and according to their linguistic understanding. 
We hope that the new platform, in combination with the methods proposed here, will allow us to label the new and larger dataset much more quickly. 

A second limitation of the new release is that the markables in the corpus were automatically extracted;
thus, the quality of the mentions is lower than in corpora in which they were hand-identified. 
    The approach followed in these years has been to ask our players to signal issues; as a result,  tens of thousands of markables were hand-corrected. 
    However, this approach doesn't really lend itself to scaling up. 
    Thus, in  our new platforms we are following a different strategy: asking our players to do the corrections themselves, by including also games to check other levels of linguistic interpretation.


\section*{Acknowledgements}

This research was supported in part by  the \ACRO{anawiki} project, funded by \ACRO{epsrc} (EP/F00575X/1)\footnote{\url{https://anawiki.essex.ac.uk/}}, in part by the {\ACRO{DALI}} project, funded by the European Research Council (\ACRO{erc}), Grant agreement ID: 695662.\footnote{\url{http://www.dali-ambiguity.org}}

\bibliography{custom}
\bibliographystyle{acl_natbib}

\newpage
\onecolumn
\appendix
\section*{Appendix}
\section{The previous release of the corpus}
\label{sec:appendix:pd2}


{\PD} 2 consisted 
of a total of 542 documents containing 408K tokens and 108K markables 
from two main genres: Wikipedia articles and fiction from the Gutenberg collection.
This version of the corpus was divided in two subsets. 
The subset referred to to  as {\PDSILVER} consisted of  
497 documents, for a total of 384K tokens and 101K markables, whose annotation was completed--i.e. 8 judgments per markable were collected, and 4 validations per interpretation--as of 12th of October 2018.
In these documents, an aggregated (`silver') label obtained through {\MPA} 
is also provided.
45 additional documents were also gold-annotated by two experts annotators.
The  subset of the corpus for which both gold and silver annotations are available was called {\PDGOLD}, as it is intended to be used as test set.\footnote{\PDGOLD is the dataset  released in 2016 as {\PD} corpus, Release 1 \cite{chamberlain-et-al:LREC16}.}
The gold subset consists of 
a total of 23K tokens and 6K markables.
The contents of the {\PD} 2 corpus are summarized in Table \ref{tab:PD2_summary}.

\begin{table}[]
    \centering
    \small
    \begin{tabular}{ccccc}
    \toprule
              & & Docs & Tokens & Markables 
              \\\midrule
    \multirow{4}{*}{\PDGOLD}   
       & Gutenberg & 5 &   7536	&   1947 (1392) \\
       & Wikipedia & 35&  15287 &	3957 (1355) \\
       & GNOME     & 5 &	989	&    274 (96) \\
       & Subtotal  & 45&  23812	&   6178 (2843) \\
       \midrule
    \multirow{4}{*}{\PDSILVER}   
       & Gutenberg & 145& 158739&  41989 (26364)\\
       & Wikipedia & 350& 218308&  57678 (19444)\\
       & Other     & 2	&   7294&	2126 (1339)\\
       & Subtotal  & 497& 384341& 101793 (47147)\\
       \midrule
    All 
       & Total     & 542& 408153& 107971 (49990)\\
       \bottomrule
    \end{tabular}
    \caption{Summary of the contents of the 2019 release of the {\PD} corpus. The numbers in parentheses indicate the total number of markables that are non-singletons.}
    \label{tab:PD2_summary}
\end{table}

\section{Detailed Evaluation Results}
\label{sec:appendix}
This appendix section includes the detailed evaluation results for this paper. More specifically,  Table \ref{tab:test-results-pred-mention-detailed} and Table \ref{tab:non-referring-results-pred-mention-detailed} show the detailed scores for our experiments on predicted mentions (discussed in Section \ref{section-test-results}); Table \ref{tab:test-results-pred-mention-domain-detailed} and Table \ref{tab:non-referring-results-pred-mention-domain-detailed} show the detailed scores of coreference and non-referring expressions for the domain specific training experiments set out in Section \ref{sec:domain-specific}. Table \ref{tab:test-results-pred-mention-longformer-detailed} shows the detailed scores for the long/short documents experiments discussed in Section \ref{section-long-docs-results}.

\begin{table*}[!htbp]
\centering
\small
\begin{tabular}{l l l l l l l l l l l l}
\toprule
\multirow{2}{*}{Singletons} & \multirow{2}{*}{Train Data} & \multicolumn{3}{l}{MUC} & \multicolumn{3}{l}{BCUB} & \multicolumn{3}{l}{CEAFE} & \multirow{2}{*}{\begin{tabular}[c]{@{}l@{}}Avg.\\ F1\end{tabular}} \\ \cmidrule{3-11}
 & & P & R & F1& P & R & F1& P& R  & F1 & \\ \midrule
\multirow{3}{*}{Included} & 
\PDTWOTRAIN&83.2 & 60.6 & 70.1&73.9 & 54.8 & 62.9&59.7 & 67.5 & 63.4&65.5\\ 
&\PDCOMTRAIN&83.1 & 62.1 & 71.1&74.7 & 54.5 & 63.0&62.4 & 66.2 & 64.3&66.1\\
&{\PDFULLTRAIN} (original)&84.2 & 58.6 & 69.1&76.0 & 52.5 & 62.1&60.2 & 66.8 & 63.4&64.9\\
&{\PDFULLTRAIN} (with Coref)& 83.4 & 63.4 & 72.0&74.5 & 55.3 & 63.5&63.4 & 66.5 & 64.9&66.8\\\midrule 
\multirow{3}{*}{Excluded} &
\PDTWOTRAIN& 83.2 & 60.6 & 70.1&72.4 & 36.6 & 48.6&52.8 & 34.9 & 42.0&53.6\\ 
& \PDCOMTRAIN&83.1 & 62.1 & 71.1&71.6 & 37.6 & 49.3&53.6 & 36.8 & 43.6&54.7\\ 
&{\PDFULLTRAIN} (original)&84.2 & 58.6 & 69.1&73.4 & 33.4 & 46.0&55.1 & 36.1 & 43.7&52.9\\
& {\PDFULLTRAIN} (with Coref)& 83.4 & 63.4 & 72.0&71.0 & 39.0 & 50.4&56.2 & 38.8 & 45.9&56.1\\ \bottomrule
\end{tabular}
\caption{The CoNLL scores for the  \newcite{yu-etal-2020-cluster} system trained on different training sets and tested on the {\PDTEST} set in predicted mention setting. }
\label{tab:test-results-pred-mention-detailed}
\end{table*}

\begin{table}[!htbp]
\centering
\small
\begin{tabular}{l l l l}
\toprule
Train data& P     & R     & F1    \\ \midrule
\PDTWOTRAIN  &73.8 & 24.6 & 36.8  \\ 
\PDCOMTRAIN &71.1 & 27.3 & 39.4 \\
\PDFULLTRAIN (original)&75.1 & 23.3 & 35.5\\
\PDFULLTRAIN (with Coref)& 77.9 & 27.0 & 40.1\\ \bottomrule
\end{tabular}
\caption{Non-referring scores for \newcite{yu-etal-2020-cluster} system trained on different training sets and tested on the {\PDTEST} set in predicted mention setting.}
\label{tab:non-referring-results-pred-mention-detailed}
\end{table}

\begin{table*}[!htbp]
\centering
\small
\begin{tabular}{l l l l l l l l l l l l}
\toprule
\multirow{2}{*}{Singletons} & \multirow{2}{*}{Train Data} & \multicolumn{3}{l}{MUC} & \multicolumn{3}{l}{BCUB} & \multicolumn{3}{l}{CEAFE} & \multirow{2}{*}{\begin{tabular}[c]{@{}l@{}}Avg.\\ F1\end{tabular}} \\ \cmidrule{3-11}
 & & P & R & F1& P & R & F1& P& R  & F1 & \\ \midrule
\multicolumn{12}{c}{Gutenberg} \\ \midrule
\multirow{2}{*}{Included} & 
\DOMAINTRAIN&87.3 & 75.2 & 80.8&71.5 & 57.6 & 63.8&61.3 & 73.4 & 66.8&70.4\\
&{\PDFULLTRAIN} (with Coref)&87.6 & 75.9 & 81.3&73.2 & 57.8 & 64.6&65.0 & 72.5 & 68.5&71.5 \\\midrule 
\multirow{2}{*}{Excluded} &
\DOMAINTRAIN&87.3 & 75.2 & 80.8&70.2 & 42.0 & 52.5&56.3 & 48.5 & 52.1&61.8\\ 
& {\PDFULLTRAIN} (with Coref)&87.6 & 75.9 & 81.3&69.9 & 42.7 & 53.0&56.3 & 48.2 & 51.9&62.1\\ \midrule
\multicolumn{12}{c}{Wikipedia} \\ \midrule
\multirow{2}{*}{Included} & 
\DOMAINTRAIN&75.4 & 52.0 & 61.6&72.8 & 54.0 & 62.0&61.2 & 62.8 & 62.0&61.9\\
&{\PDFULLTRAIN} (with Coref)&78.1 & 51.3 & 61.9&75.5 & 53.3 & 62.5&62.4 & 62.7 & 62.5&62.3 \\\midrule 
\multirow{2}{*}{Excluded} &
\DOMAINTRAIN&75.4 & 52.0 & 61.6&69.8 & 36.9 & 48.3&54.1 & 35.5 & 42.9&50.9\\ 
& {\PDFULLTRAIN} (with Coref)&78.1 & 51.3 & 61.9&72.3 & 35.8 & 47.9&56.2 & 33.5 & 42.0&50.6\\ 
\bottomrule
\end{tabular}
\caption{The CoNLL scores for \newcite{yu-etal-2020-cluster} system trained on different training sets and tested on the {\PDTEST} set of different domains in predicted mention setting. }
\label{tab:test-results-pred-mention-domain-detailed}
\end{table*}

\begin{table}[!htbp]
\centering
\small
\begin{tabular}{l l l l}
\toprule
Train data& P     & R     & F1    \\ \midrule
\multicolumn{4}{c}{Gutenberg}\\\midrule
\DOMAINTRAIN  & 79.9 & 30.2 & 43.8 \\ 
{\PDFULLTRAIN} (with Coref) & 84.0 & 30.6 & 44.9\\ \midrule
\multicolumn{4}{c}{Wikipedia}\\\midrule
\DOMAINTRAIN  & 71.6 & 24.1 & 36.1\\ 
{\PDFULLTRAIN} (with Coref) &72.4 & 24.0 & 36.0 \\ 
\bottomrule
\end{tabular}
\caption{Non-referring scores for \newcite{yu-etal-2020-cluster} system trained on different training sets and tested on the {\PDTEST} set of different domains in predicted mention setting.}
\label{tab:non-referring-results-pred-mention-domain-detailed}
\end{table}

\begin{table*}[!htbp]
\centering
\small
\begin{tabular}{l l l l l l l l l l l l}
\toprule
\multirow{2}{*}{Settings} & \multirow{2}{*}{Model} & \multicolumn{3}{l}{MUC} & \multicolumn{3}{l}{BCUB} & \multicolumn{3}{l}{CEAFE} & \multirow{2}{*}{\begin{tabular}[c]{@{}l@{}}Avg.\\ F1\end{tabular}} \\ \cmidrule{3-11}
 & & P & R & F1& P & R & F1& P& R  & F1 & \\ \midrule
\multirow{2}{*}{Short Doc} & 
\LONGFORMER&96.2&61.5&75.0 &88.4&45.9&60.4 &74.8&34.7&47.4 &61.0\\
&\ROBERTA&96.3&60.8&74.5 &89.3&45.1&59.9 &71.1&33.9&45.9 &60.1\\ 
\midrule
\multirow{2}{*}{Long Doc} & 
\LONGFORMER &94.2&71.6&81.3 &77.6&53.3&63.1 &73.2&46.7&57.0 &67.2\\
&\ROBERTA &94.4&71.1&81.1 &76.3&49.4&59.9 &71.9&43.8&54.4 &65.2\\ 
\midrule

\multirow{2}{*}{All Doc} & 
\LONGFORMER &94.9&67.5&78.9 &81.6&50.2&62.2 &73.8&41.3&52.9 &64.7\\
&\ROBERTA &95.1&66.9&78.5 &81.1&47.6&60.0 &71.6&39.3&50.7 &63.1\\ 
\bottomrule
\end{tabular}
\caption{The CoNLL scores for {\LONGFORMER} and {\ROBERTA} systems trained on {\PDFULLTRAIN} and tested on the {\PDTEST} set using gold mentions in a singleton excluded setting.}
\label{tab:test-results-pred-mention-longformer-detailed}
\end{table*}

\end{document}